\documentclass[letterpaper, 10 pt, conference]{ieeeconf}  

\IEEEoverridecommandlockouts                              

\usepackage{graphics} 
\usepackage{epsfig} 
\usepackage{times} 
\usepackage{amsmath} 
\usepackage{amssymb}  

\usepackage{booktabs}   
\usepackage{hyperref}

\usepackage[dvipsnames]{xcolor}

\title{\LARGE \bf
High-Bandwidth Tactile-Reactive Control for Grasp Adjustment
}

\author{Yonghyeon Lee$^{1*}$, Tzu-Yuan Lin$^{1*}$, Alexander Alexiev$^{1}$, and Sangbae Kim$^{1}$
\thanks{*Equal contribution.}
\thanks{$^{1}$The authors are with the Department of Mechanical Engineering, Massachusetts Institute of Technology, Cambridge, MA 02139 USA (e-mail: yhl@mit.edu; tzuyuan@mit.edu; sangbae@mit.edu).}
}

\begin{document}

\maketitle
\thispagestyle{empty}
\pagestyle{empty}

\begin{abstract}
 Vision-only grasping systems are fundamentally constrained by calibration errors, sensor noise, and grasp-pose prediction inaccuracies, leading to unavoidable contact uncertainty in the final stage of grasping.
 High‑bandwidth tactile feedback, when paired with a well‑designed tactile‑reactive controller, can significantly improve robustness in the presence of perception errors. 
 This paper contributes to controller design by proposing a purely tactile-feedback grasp-adjustment algorithm. The proposed controller requires neither prior knowledge of the object’s geometry nor an accurate grasp pose, and is capable of refining a grasp even when starting from a crude, imprecise initial configuration and uncertain contact points. Through simulation studies and real‑world experiments on a 15‑DoF arm–hand system (featuring an 8‑DoF hand) equipped with fingertip tactile sensors operating at 200~Hz, we demonstrate that our tactile-reactive grasping framework effectively improves grasp stability. 
 Project page: \href{https://reflexivegrasp.github.io}{https://reflexivegrasp.github.io}.
\end{abstract}
\section{Introduction}
A typical grasping pipeline relies on visual inputs, such as RGB images or point clouds, to estimate a target grasp pose and then guides the robot to approach and grasp the object. A substantial body of recent work has focused on enhancing the visual perception component, particularly grasp pose generation and object geometry reconstruction, leading to promising results for handling novel, unseen objects, including but not limited to~\cite{mahler2017dex,kim2022dsqnet,lee2023nfl,t2sqnet,fang2023anygrasp,lim2024equigraspflow,sundermeyer2021contact,wen2023bundlesdf,millane2024nvblox}.

However, systems that rely purely on vision are inevitably susceptible to errors in the perception pipeline, such as calibration inaccuracies, sensor noise, grasp pose prediction errors, and geometry reconstruction errors. Many of these errors are unavoidable and can be particularly severe due to occlusions, which frequently arise from the robot’s own hand or the object itself.
These issues often lead to uncertain contacts between the gripper fingertips and objects during the final stages of grasping, occasionally resulting in grasp failure. 

Referred to as the {\it last centimeter problem} in~\cite{saloutos2023towards}, this challenge highlights the sharp rise in complexity resulting from unpredictable contact dynamics in the final phase of grasping. The authors addressed it by leveraging a high-bandwidth tactile sensor to enable fast closed-loop feedback control -- referred to as a {\it reflexive control} -- that locally refines the grasp pose toward a more stable configuration, without relying on vision data in the loop. Inspired by this approach, we aim to extend it for broader applicability in more general scenarios involving arbitrary object shapes and grasping directions, whereas the existing method is limited to planar grasping and assumes circular horizontal cross-sections.

\begin{figure}[!t]
    \centering
    \includegraphics[width=1\linewidth]{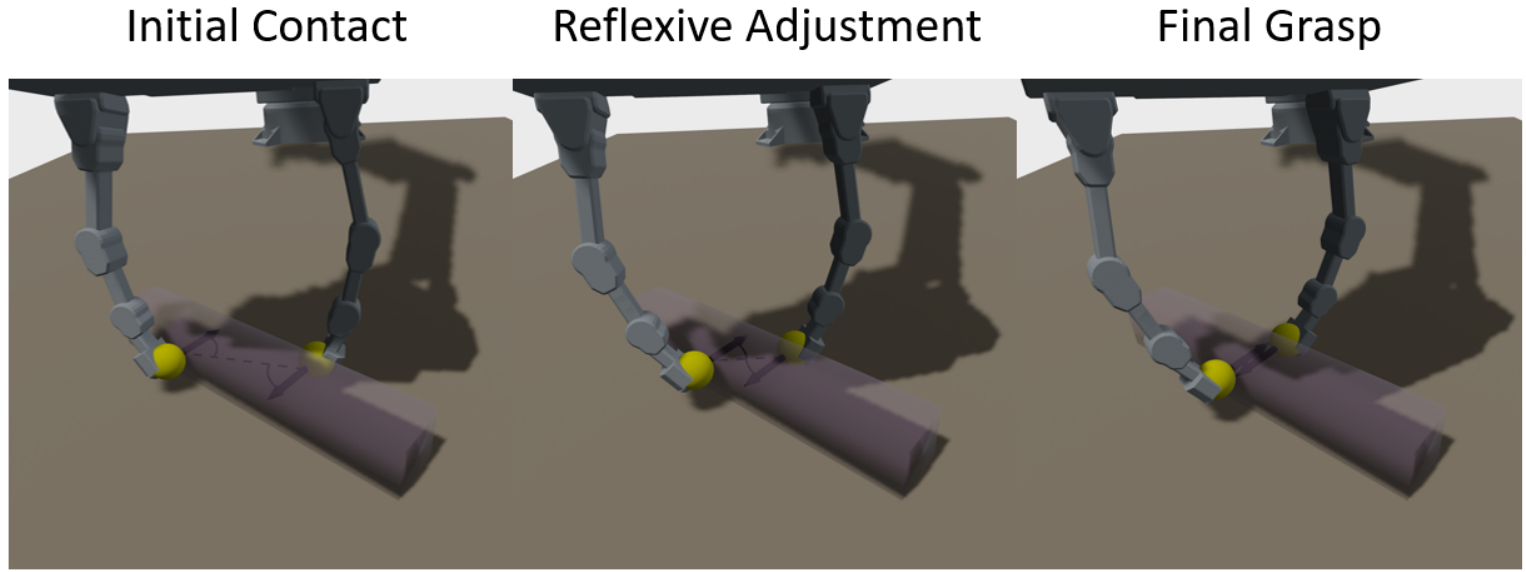}
    \caption{A representative example of our reflexive grasping controller starting from an imprecise initial gripper configuration. The dashed line connects the two contact points, the black arrows indicate the contact normal vectors, and the arcs represent the angles between them -- smaller angles indicate more stable grasps. }
    \label{fig:intro}
\end{figure}

Throughout this paper, we focus on the gripper-closing stage, assuming the robot has already reached a state where it is ready to close the gripper (any reaching policy may be used to arrive at this initial configuration). Our goal is to develop a tactile-reactive controller -- hereafter referred to as {\it reflexive controller} -- that adjusts the gripper to a stable grasp using only fingertip tactile sensors that provide contact positions. The algorithm is designed to operate from crude, imprecise pre-grasp configurations, as long as the fingertips are likely to make contact with the object upon closing, as illustrated in Fig.~\ref{fig:intro}.

\begin{figure*}[!t]
    \centering
    \includegraphics[width=1\linewidth]{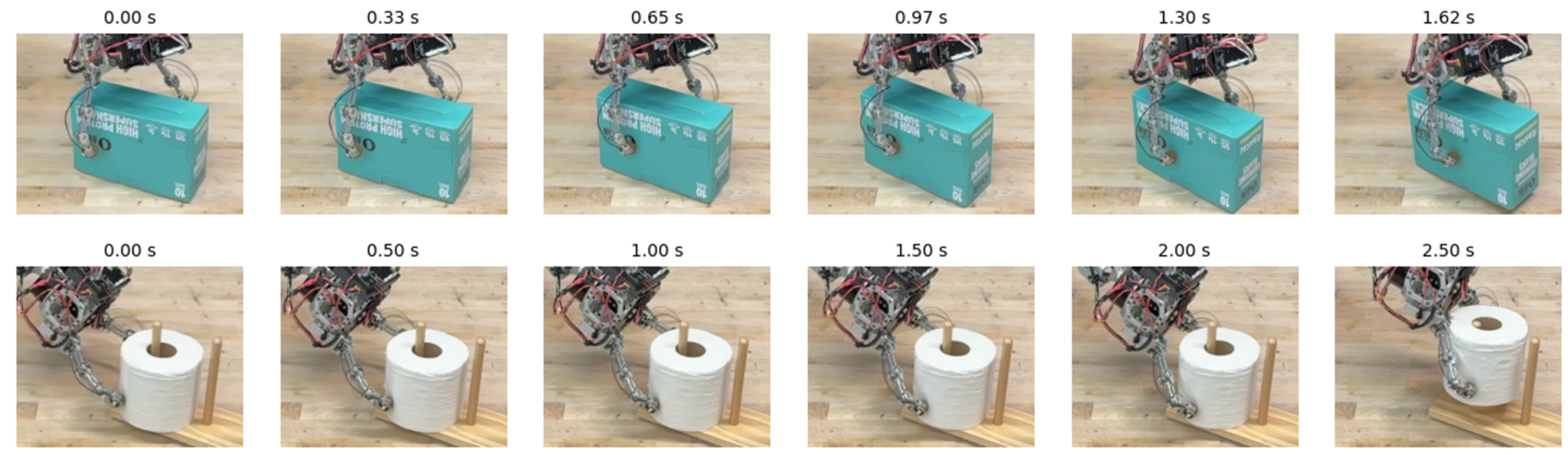}
    \caption{Real-world examples of our reflexive grasping controller starting from an imprecise initial gripper configuration, using a 15-DoF arm-hand system (7-DoF arm and 8-DoF hand) on a manipulation platform adapted from~\cite{saloutos2023towards}. The fingertip tactile sensor outputs contact points and surface normals at 200 Hz, while the tactile-reactive feedback controller computes tangent-space motions that guide the fingertips toward a more stable, antipodal grasp in real time.}
    \label{fig:intro2}
\end{figure*}

The proposed algorithm assumes no prior knowledge of object geometry and requires no grasp-pose generation. It relies solely on tactile sensing, using contact points and their associated normals computed from the fingertip surface geometry. 
We begin by formulating a grasp stability objective based on contact positions and normals. Then the desired 6-DoF fingertip velocities are computed to minimize this objective while remaining within the contact tangent spaces, ensuring each fingertip moves orthogonally to its contact normal for local readjustment.
Finally, the corresponding joint velocities are obtained by solving a joint-space Quadratic Program (QP) that incorporates additional constraints, including self-collision avoidance, environmental collisions, joint limits, and velocity bounds.

We present a case study using a two-finger, 8-DoF gripper mounted on a 7-DoF arm, adapted from~\cite{saloutos2023towards}. To evaluate grasp stability, we formulate an antipodal grasp stability function~\cite{nguyen1988constructing,murray1994mathematical} and derive a desired fingertip linear velocity from its gradient. We propose and compare two approaches for determining the gradient direction: (i) a straightforward method using Projected Gradient Descent (PGD), and (ii) a novel Cross-Finger Gradient Descent (CFGD), in which each finger moves to minimize the angle induced by the other finger. Our simulation results show that while PGD frequently converges to local minima, CFGD consistently achieves better performance across all tested cases.

Real-robot experiments are conducted on a system equipped with spherical fingertip tactile sensors~\cite{saloutos2023design,saloutos2023towards}, which provide contact probabilities and contact positions at 200 Hz. 
In experimental settings where the tactile sensors provide sufficiently reliable measurements, the proposed method achieves successful reflexive grasping; representative trials are shown in Fig.~\ref{fig:intro2}.
\section{Related Work: Grasp Adjustment with Tactile Feedback}
Typically, an autonomous grasp consists of two stages: grasp planning and grasp execution.  
In practice, inevitable grasp–pose estimation errors -- such as those caused by occlusions or calibration inaccuracies -- and additional tracking errors in control lead to deviations from the desired grasp and potential grasp failure.  
To address this issue, prior work has explored grasp adjustment during the execution phase, relying on local tactile measurements and employing a variety of hand designs, tactile sensors, and algorithms under various assumptions~\cite{dang2013grasp,chebotar2016self,hogan2018tactile,liu2022multi,saloutos2022fast,saloutos2023towards}.

Some studies rely on tactile databases or adopt data‑driven, learning‑based approaches. A tactile experience database constructed from shape primitives is leveraged~\cite{dang2013grasp}, where local geometric similarities enable the transfer of tactile feedback from known primitives to novel objects. A self‑supervised framework is introduced~\cite{chebotar2016self}, in which spatio‑temporal tactile features allow early grasp‑stability prediction and guide a reinforcement‑learning policy for regrasping. A tactile‑based regrasp policy is proposed~\cite{hogan2018tactile} that builds on a learned tactile‑based grasp‑quality metric and a model that simulates local transformations of tactile imprints to search for grasp adjustments that improve stability.

In contrast, our method does not rely on datasets or learning, but instead adjusts the grasp pose directly from local tactile feedback, given a specified grasp-stability criterion, similar in spirit to~\cite{liu2022multi,saloutos2022fast,saloutos2023towards}. While~\cite{saloutos2022fast,saloutos2023towards} assume specific object geometries (circular horizontal cross-sections) and focus primarily on planar grasping, the work most closely related to ours is~\cite{liu2022multi}, which makes no prior assumptions about object geometry or grasping direction. In~\cite{liu2022multi}, a tactile Jacobian -- chosen via a heuristic strategy -- maps deviations in contact features (e.g., contact normal forces, contact positions) to fingertip sliding or rotation, while a multi-finger inverse kinematics module computes coordinated palm and finger motions without breaking contact. This approach is conceptually similar to ours at a high level. A key limitation, however, is that the desired contact features are defined independently for each fingertip, making it unsuitable for cases where stability is determined by geometric relationships among multiple fingertips (e.g., a two-finger antipodal grasp).

Compared with these existing works, our framework offers several unique features:  
(i) it provides a principled and general formulation for multi‑DoF, multi‑finger hands based on a differentiable grasp‑stability function;  
(ii) it makes no assumptions about the geometry of objects;  
(iii) it relies solely on contact position information; and  
(iv) it incorporates full‑body (typically arm–hand) control algorithms that explicitly consider joint‑level constraints.
\section{A Hierarchical Reflexive Control Framework for Grasp Adjustment}
We begin this section by introducing the key assumptions and notation used throughout the paper. We consider a fully actuated robotic system composed of an articulated arm and a multi-fingered gripper with $m$ fingertips and a total of $n$ degrees of freedom. The system's joint configuration is represented by $q \in \mathbb{R}^n$. Each fingertip $i \in \{1, \dots, m\}$ is associated with a 3D position $x_i \in \mathbb{R}^3$ and a rotation matrix $R_i \in \mathbb{R}^{3 \times 3}$, defining its pose. 

The linear and angular velocities of each fingertip are related to joint velocities via geometric Jacobians $J_i^x(q) \in \mathbb{R}^{3 \times n}$ and $J_i^R(q) \in \mathbb{R}^{3 \times n}$, satisfying:
$$
\dot{x}_i = J_i^x(q) \dot{q}, \quad \dot{R}_i R_i^\top = [J_i^R(q)\, \dot{q}],
$$
where $[\omega]$ denotes the skew-symmetric matrix corresponding to an angular velocity vector $\omega \in \mathbb{R}^3$, such that $[\omega]v = \omega \times v$ for any $v \in \mathbb{R}^3$~\cite{lynch2017modern}.

\begin{figure}[!t]
    \centering
    \includegraphics[width=0.85\linewidth]{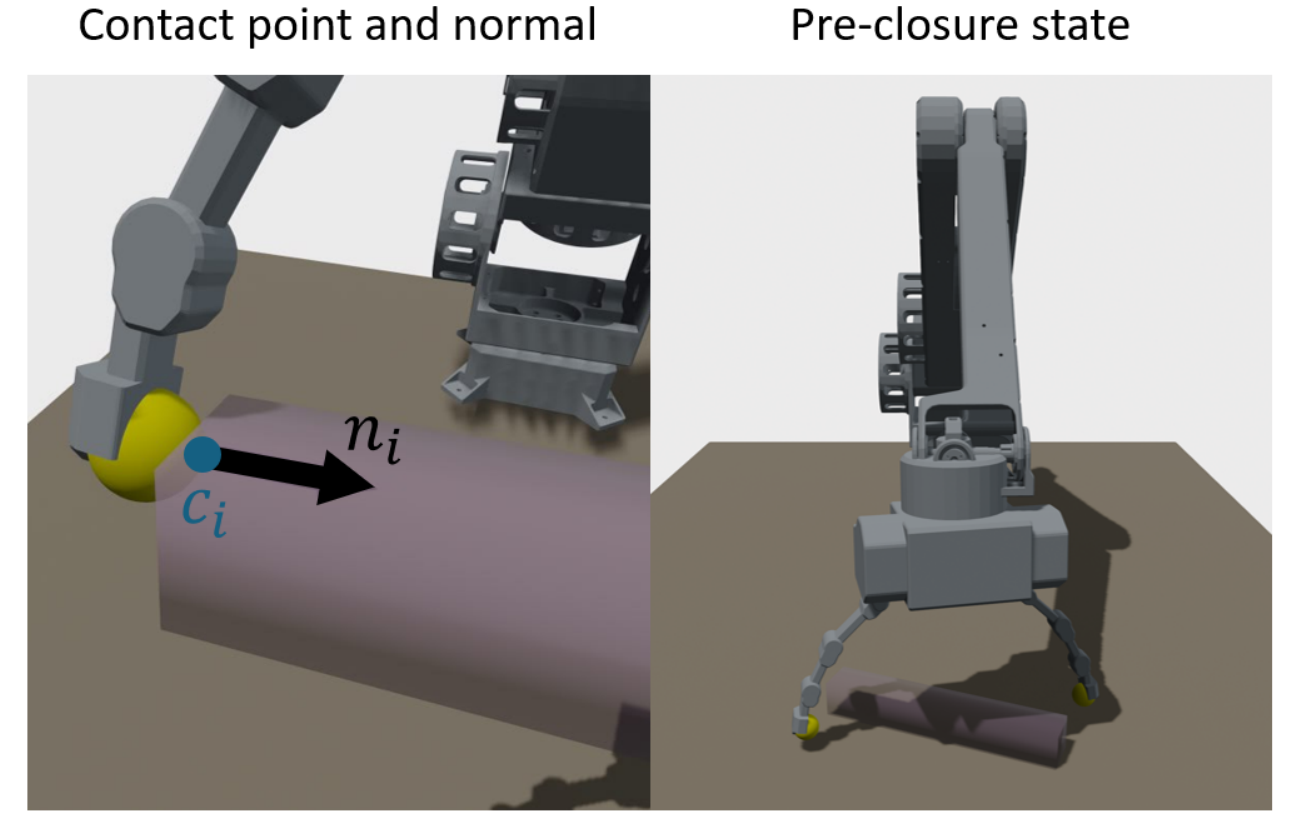}
    \caption{{\it Left}: The fingertip is covered with a yellow surface that enables contact position sensing. The contact point is denoted by $c_i$, and the corresponding contact normal by $n_i$; in this example, $n_i$ is determined from $c_i$ based on the known spherical fingertip geometry. {\it Right}: An example of a pre-closure state, from which closing the gripper will cause the fingertips to make contact with the object, although the resulting grasp may be unstable.}
    \label{fig:example_contact_surface}
\end{figure}

Each fingertip is covered with a surface that enables contact position sensing, e.g., using an elastomer layer such as rubber~\cite{saloutos2023towards, saloutos2022fast}. We assume point contact, where each fingertip touches the object at a single point, estimated as $c_i \in \mathbb{R}^3$ for $i = 1, \ldots, m$. The corresponding contact normal $n_i \in \mathbb{R}^3$ is computed from the analytical geometry of the fingertip surface and points outward. For smooth object boundaries, this normal aligns with the object’s surface normal. For example, see Fig.~\ref{fig:example_contact_surface} ({\it Left}).

We assume the existence of a reaching policy (which can be arbitrary) that brings the gripper to a pre-closure state, such that closing the gripper -- e.g., moving each fingertip toward the center of the fingertip positions -- results in contact between all fingertips and the object boundary. For example, see Fig.~\ref{fig:example_contact_surface} ({\it Right}). The goal of the method proposed in this section is to refine the fingertip positions and orientations based on high-bandwidth tactile feedback, adjusting them toward a stable grasp configuration.

Our framework adopts a hierarchical structure consisting of two main steps: (i) constructing desired fingertip 6-DoF velocity fields conditioned on tactile feedback, and (ii) solving a joint-space Quadratic Program (QP) to compute joint velocities that track the desired fingertip motions while satisfying joint-level constraints such as joint limits and collision avoidance. 
The overall control frequency is bounded by the tactile sensing feedback rate, as both the velocity field construction -- based on analytical expressions -- and the QP solving can be executed significantly faster.
Each step is detailed in the following sections.

\subsection{Construction of Desired Fingertip Velocity Fields}
We define two discrete control modes in our reflexive regrasp controller: (i) gripper-closing, and (ii) grasp adjustment. These modes alternate until a stable grasp configuration is reached. Grasp stability is evaluated using a criterion defined as a function of the contact positions and normals:
\begin{equation}
\label{eq:gso}
    f(c_1, \ldots, c_m, n_1, \ldots, n_m),
\end{equation}
where a lower value of $f$ indicates a more stable grasp. A grasp is considered stable if the function value falls below a predefined threshold and contact is detected at all fingertips.

Initially, the controller operates in gripper-closing mode when no contact is detected at any fingertip. Once contact is detected at all fingertips and the grasp is deemed unstable, it transitions to grasp adjustment mode. If any fingertip loses contact during this process, the controller reverts to gripper-closing mode. These two modes alternate until a stable grasp is achieved, at which point the control process terminates.

In the gripper-closing mode, the desired fingertip linear velocities are defined as
\begin{equation}
\label{eq:gc_lftv}
    \dot{x}_{i,\mathrm{des}} = V_{\mathrm{c}} \frac{x_c - x_i}{\|x_c - x_i\|}, \quad i = 1, \ldots, m,
\end{equation}
where $x_c = \tfrac{1}{m}\sum_{i=1}^{m} x_i$ is the centroid of the fingertip positions, and $V_{\mathrm{c}}$ is a constant speed parameter.  
This formulation naturally drives the fingertips inward, allowing the gripper to close around the object and make contact.  
Once contact is detected, the latest contact normal directions are retained and used to define the desired gripper-closing velocity as  
\begin{equation}
\label{eq:gc_lftv2}
    \dot{x}_{i,\mathrm{des}} = V_{\mathrm{c}}\, n_i, \quad i = 1, \ldots, m.
\end{equation}

In the grasp adjustment mode, the desired linear velocities are defined to move the fingertip in a direction that decreases the grasp stability objective $f$ in~(\ref{eq:gso}). In this process, to avoid excessive pushing against the object, the desired velocity direction is constrained to lie within the contact tangent space -- i.e., it satisfies $\dot{x}_{i,\mathrm{des}} \perp n_i$, where $n_i$ is the contact normal at fingertip $i$.

\begin{figure}[!t]
    \centering
    \includegraphics[width=1\linewidth]{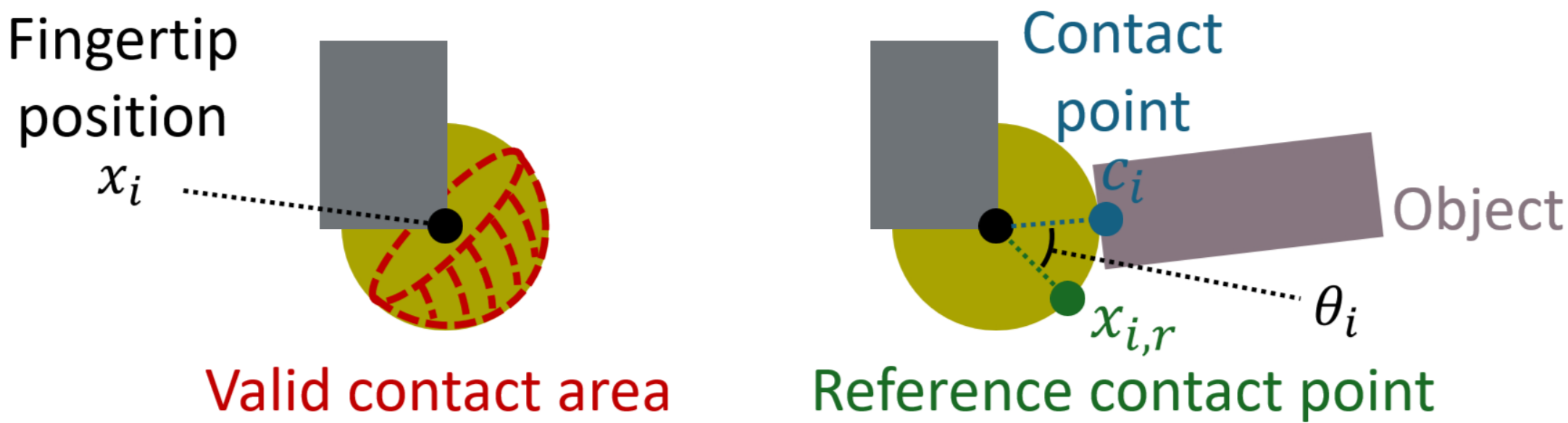}
    \caption{An illustrative example of a fingertip showing the valid contact area (red shaded region, {\it Left}) and the reference contact point (green point, {\it Right}). The contact point (blue, {\it Right}) lies within the valid contact area if and only if the angle $\theta_i$ is less than $\pi/2$.}
    \label{fig:example_orientation}
\end{figure}

One straightforward approach is to use the Projected Gradient Descent (PGD) direction, that is,
\begin{equation}
\label{eq:ga_lftv}
    \dot{x}_{i,\mathrm{des}} = - V_a \left( I - n_i n_i^\top \right) \nabla_{c_i} f,
\end{equation}
where $V_a$ is the constant speed parameter, \( \nabla_{c_i} f \) is the gradient of the grasp stability objective with respect to the contact position $c_i$, and the projection matrix \( I - n_i n_i^\top \) ensures that the motion remains within the contact tangent space.
While PGD is the most natural baseline to try, its effectiveness depends heavily on the choice of grasp stability function $f$, and in some cases, more suitable alternatives exist. In a later case study with a two-finger gripper, we show that PGD is often trapped in local minima and introduce an alternative approach that achieves improved convergence.

Additionally, we construct the desired fingertip angular velocities $\omega_{i,\mathrm{des}}$ to rotate each fingertip such that the contact point lies within a valid contact area on the fingertip surface -- i.e., a region where tactile signals can be reliably estimated, e.g., see Fig.~\ref{fig:example_orientation} (\textit{Left}). 
Specifically, we define a reference contact point $x_{i,r}$ on each fingertip surface, which is the central point of the contact surface as shown in Fig.~\ref{fig:example_orientation} (\textit{Right}).
To ensure that the contact point lies within the valid region, we require that the angle $\theta_i$ between the vectors $x_{i,r} - x_i$ and $c_i - x_i$ (where $c_i$ is the contact point) remains below a specified threshold $\delta$; see Fig.~\ref{fig:example_orientation} ({\it Right}). 

We note that $x_{i,r} - x_i$ depends on the fingertip orientation $R_i$; for example, it can be written as a weighted sum of the column vectors of $R_i$. Intuitively, one may view $x_{i,r} - x_i$ as a vector originating at $x_i$ (the center of the sphere), while $R_i$ provides three orthonormal basis vectors anchored at the same point.
Accordingly, $\theta_i$ is also a function of $R_i$, given a current contact position $c_i$ and fingertip position $x_i$.

When $\theta_i$ is higher than the threshold value, we need to rotate the fingertip to reduce it. Accordingly, we define the desired angular velocity $\omega_{i,\mathrm{des}}$ as the negative gradient direction as follows:
\begin{equation}
\label{eq:ga_aftv}
    \omega_{i, {\rm des}} = - W  \big[ \frac{\partial \theta_i}{\partial R_i} R_i^T \big]_{\rm vec},
\end{equation}
where $W$ is a constant rotational speed parameter. 
The operator $[\cdot]_{\mathrm{vec}}$ maps a $3 \times 3$ skew-symmetric matrix to its corresponding vector -- i.e., for any vector $v \in \mathbb{R}^3$, if $S$ is skew-symmetric, then $Sv = [S]_{\mathrm{vec}} \times v$.
We only track this angular velocity if $\theta_i > \delta$; otherwise, tracking is unnecessary. This is implemented by controlling a weight parameter in the objective function of the QP problem, as discussed in the next section.

\subsection{Tracking Fingertip Velocities via Joint-Space QP}
Desired joint velocities are obtained by solving a joint-space Quadratic Program (QP) formulated to track the specified fingertip velocities. In the gripper-closing mode, the QP follows the linear fingertip velocity defined in~(\ref{eq:gc_lftv}), and switches to~(\ref{eq:gc_lftv2}) once the first contact is established. In the grasp adjustment mode, it tracks both the linear fingertip velocity~(\ref{eq:ga_lftv}) and the angular fingertip velocity~(\ref{eq:ga_aftv}), while satisfying all joint-level constraints.

Let \( \Gamma(q) \in \mathbb{R}^k \) denote the signed distance values corresponding to self-collisions and environment obstacles, computed over \( k \) selected collision pairs. Joint position limits are given by \( q_{\mathrm{min}} \) and \( q_{\mathrm{max}} \), and joint velocity limits by \( \dot{q}_{\mathrm{min}} \) and \( \dot{q}_{\mathrm{max}} \).
We assume a constant desired joint velocity \( \dot{q}_{\mathrm{des}} \) over a finite planning horizon, a positive scalar value \( H > 0 \), and approximate the future joint configuration as \( q + \dot{q}_{\mathrm{des}} \; H \). 
Drawing inspiration from prior work on real-time inverse kinematics via quadratic programming~\cite{mirrazavi2018unified, koptev2021real}, we formulate a QP that tracks the desired fingertip velocities by linearizing the constraints around the current configuration \( q \):
\begin{align}
\label{eq:qp_formulation}
    \dot{q}_{\mathrm{des}} = \arg \min_{\dot{q}} \ \ & 
    \sum_{i=1}^{m} \left\| \dot{x}_{i, \mathrm{des}} - J_i^x(q) \dot{q} \right\|^2 \nonumber \\
    & + \alpha_i \left\| \omega_{i, \mathrm{des}} - J_i^R(q) \dot{q} \right\|^2 \nonumber \\
    \noalign{\vskip 1.0ex}
    \text{subject to} \quad & \Gamma(q) + \frac{d\Gamma(q)}{dq}  \dot{q}  H \geq \epsilon_{\Gamma}, \nonumber \\ 
    & q_{\mathrm{min}} \leq q + \dot{q} H \leq q_{\mathrm{max}}, \nonumber \\
    & \dot{q}_{\mathrm{min}} \leq \dot{q} \leq \dot{q}_{\mathrm{max}},
\end{align}
where \( \alpha_i = 1 \) only if we are in grasp adjustment mode and $\theta_i > \delta$; otherwise, $\alpha_i = 0$. The parameter \( \epsilon_\Gamma > 0 \) represents the minimum safety margin, and together with the planning horizon \( H \), it determines the conservativeness of the controller’s response.

The joint-space QP can be solved very efficiently using modern optimization libraries, making the tactile sensing rate the primary bottleneck for the control frequency. Specifically, we use \texttt{Pinocchio}~\cite{carpentier2019pinocchio} for computing forward kinematics and Jacobians, and \texttt{FCL}~\cite{pan2012fcl} for computing distances and nearest points between collision geometries. The QP is solved using \texttt{OSQP}~\cite{osqp}. These modern libraries enable fast and reliable computation, allowing real-time execution of our control pipeline.

\section{Case Study: A Two-Finger Multi-DoF Gripper}
In this section, we present a case study using a two-finger multi-DoF gripper (8-DoF) mounted on a 7-DoF arm, adapted from~\cite{saloutos2023towards}. We begin by introducing a grasp stability function $f$, defined in~(\ref{eq:gso}) for a two-finger gripper, and then propose a more effective linear fingertip velocity direction for the grasp adjustment mode -- compared to the Projected Gradient Descent (PGD) direction in~(\ref{eq:ga_lftv}) -- that more efficiently reduces the value of $f$. We then conduct simulation studies to demonstrate the effectiveness of our reflexive controller relative to grasping without reflexes. Finally, we validate our approach through real-world hardware experiments.

\subsection{Antipodal Grasp Stability Function}

\begin{figure}[!t]
    \centering
    \includegraphics[width=1\linewidth]{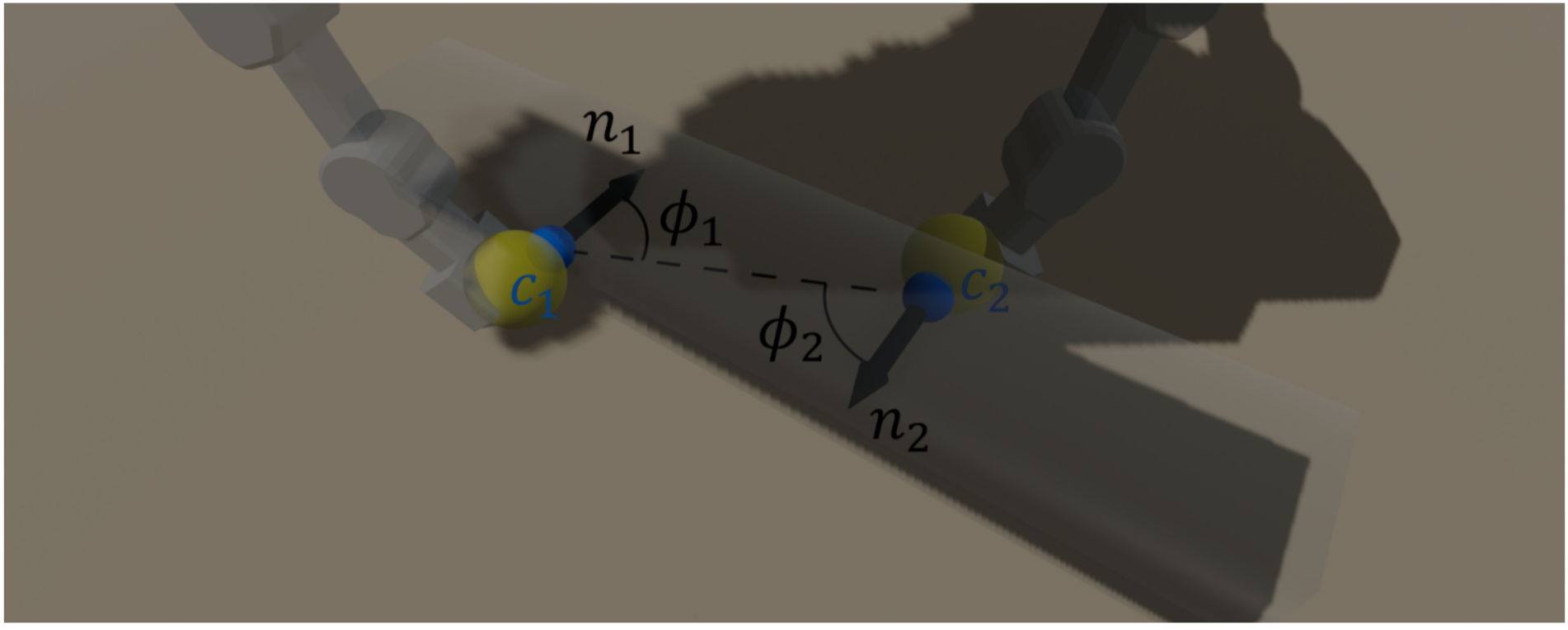}
    \caption{Two fingertips are in contact with the object; $c_1$ and $c_2$ denote the contact points, and $n_1$ and $n_2$ are the corresponding contact normals -- defined by the fingertip surface geometry. $\phi_1$ and $\phi_2$ are the stability angles, where smaller values indicate more antipodal and thus more stable grasps.}
    \label{fig:two_finger_f}
\end{figure}

We define the grasp stability function $f$ as the sum of the angles $\phi_1$ and $\phi_2$ shown in Fig.~\ref{fig:two_finger_f}:
\begin{equation}
\label{eq:antipodal_grasp_stability}
    f(c_1, c_2, n_1, n_2) = \phi_1(c_1, c_2, n_1) + \phi_2(c_1, c_2, n_2),
\end{equation}
where $\phi_1(c_1, c_2, n_1) = \angle(n_1, c_2 - c_1)$ and $\phi_2(c_1, c_2, n_2) = \angle(n_2, c_1 - c_2)$, and the angle operator is defined as
\[
\angle(a, b) := \arccos\left( \frac{a^\top b}{\|a\| \cdot \|b\|} \right).
\]
A value of $f = 0$ indicates that the two fingertips satisfy the antipodal grasp condition, allowing the object to be held securely through force balance alone. This represents the most basic type of stable grasp for two-finger grasps~\cite{nguyen1988constructing, murray1994mathematical}.

\subsection{Projected Gradient vs Cross-Finger Gradient}

\begin{figure}[!t]
    \centering
    \includegraphics[width=1\linewidth]{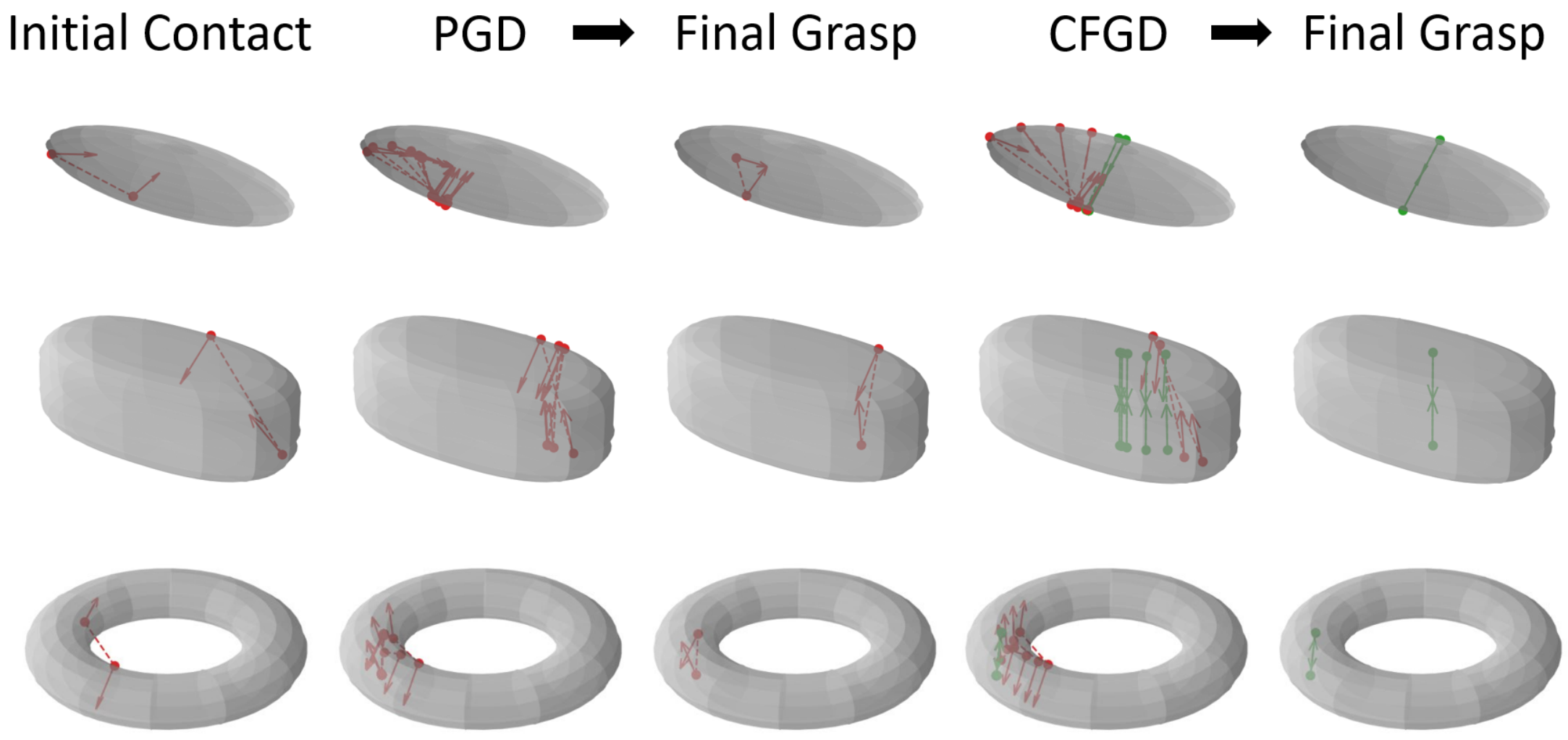}
    \caption{We iteratively move the two contact points in the directions given by Projected Gradient Descent (PGD) or Cross‑Finger Gradient Descent (CFGD), projecting them onto the object boundaries at each step from arbitrary initial contact points, until convergence. Arrows indicate the contact normals, and dashed lines connect the two contact points. Green arrows indicate that the grasp is close to an antipodal grasp. While CFGD converges to a stable grasp, PGD tends to become trapped in local minima.}
    \label{fig:CFGD}
\end{figure}

\begin{table}[!t]
\centering
\caption{Convergence rates (\%) of PGD and CFGD on different object shapes, computed over 100 random initializations.}
\label{tab:pgd_cfgd_results}
\begin{tabular}{lccc}
\toprule
 & Ellipsoid & Superquadrics & Torus \\
\midrule
PGD  & 1 \%   & 12 \%   & 2 \%   \\
CFGD & 100 \% & 99 \% & 99 \% \\
\bottomrule
\end{tabular}
\end{table}

\begin{figure}[!t]
    \centering
    \includegraphics[width=1\linewidth]{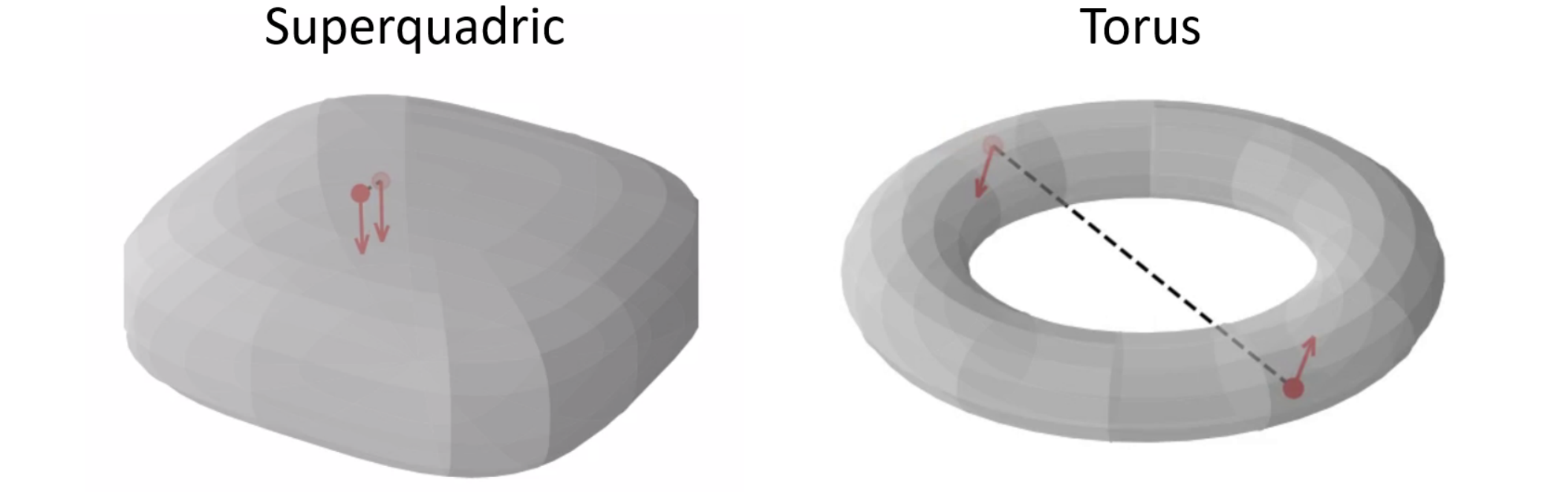}
    \caption{Antipodal grasp convergence failure cases of the CFGD method (when the angles between the dashed line and both contact normals are~$\pi/2$).
}
    \label{fig:failure_cases}
\end{figure}

For the antipodal grasp stability function~(\ref{eq:antipodal_grasp_stability}), we found that the Projected Gradient Descent (PGD) in~(\ref{eq:ga_lftv}) can often cause the fingertips to become trapped in local minima, as shown in Fig.~\ref{fig:CFGD}. 
The gradients of $\phi_1$ and $\phi_2$ with respect to each contact point $c_i$ cancel each other, so that the gradient of $f$ becomes zero even though the gradient of each angle is nonzero; that is,
\[
\nabla_{c_i} f = 0 \quad \text{because} \quad \nabla_{c_i}\phi_1 = -\,\nabla_{c_i}\phi_2 \neq 0.
\]
This phenomenon is observed even for simple convex objects and, in fact, occurs quite frequently, as shown in Fig.~\ref{fig:CFGD}.

Instead, we propose a Cross‑Finger Gradient Descent (CFGD), in which each contact point moves so as to decrease the angle defined by the other contact point; that is,
\begin{align}
\label{eq:CFGD}
    \dot{x}_{1,{\rm des}} &= - V_a \, (I - n_1 n_1^T)\,\nabla_{c_1} \phi_2(c_1, c_2, n_1), \nonumber \\
    \dot{x}_{2,{\rm des}} &= -V_a \, (I - n_2 n_2^T) \,\nabla_{c_2} \phi_1(c_1, c_2, n_2).
\end{align}
We empirically found that CFGD converges to an antipodal grasp in almost all tested cases, even for complex non‑convex shapes such as a torus, regardless of the initial contact positions, as shown in Fig.~\ref{fig:CFGD}. 

Table~\ref{tab:pgd_cfgd_results} shows the antipodal grasp convergence rates of PGD and CFGD on ellipsoids, superquadrics, and torus shapes, with randomly initialized shape parameters -- such as axis lengths for ellipsoids, size and shape parameters for superquadrics, and inner and outer radii for tori -- and random initial surface points (100 random initializations). PGD almost never converges except for a few rare cases with favorable initialization, whereas our CFGD method achieves very high convergence rates. A few failure cases of CFGD are observed when, during the update, the two contact points happen to satisfy $\phi_1 = \phi_2 = \pi/2$, as shown in Fig.~\ref{fig:failure_cases}.  

\subsection{Simulation Studies}
In this section, we present simulation studies that compare grasping without reflex to our reflexive grasping controllers using the PGD and CFGD methods. We use the MuJoCo physics engine~\cite{todorov2012mujoco}, running the low‑level controller (i.e., velocity tracking) at 1000~Hz and solving the QP (to obtain the desired joint velocities) at up to 200~Hz in a separate thread.
We test on box, cylinder, and ellipsoid objects, each with five different sizes. 

\begin{figure}[!t]
    \centering
    \includegraphics[width=\linewidth]{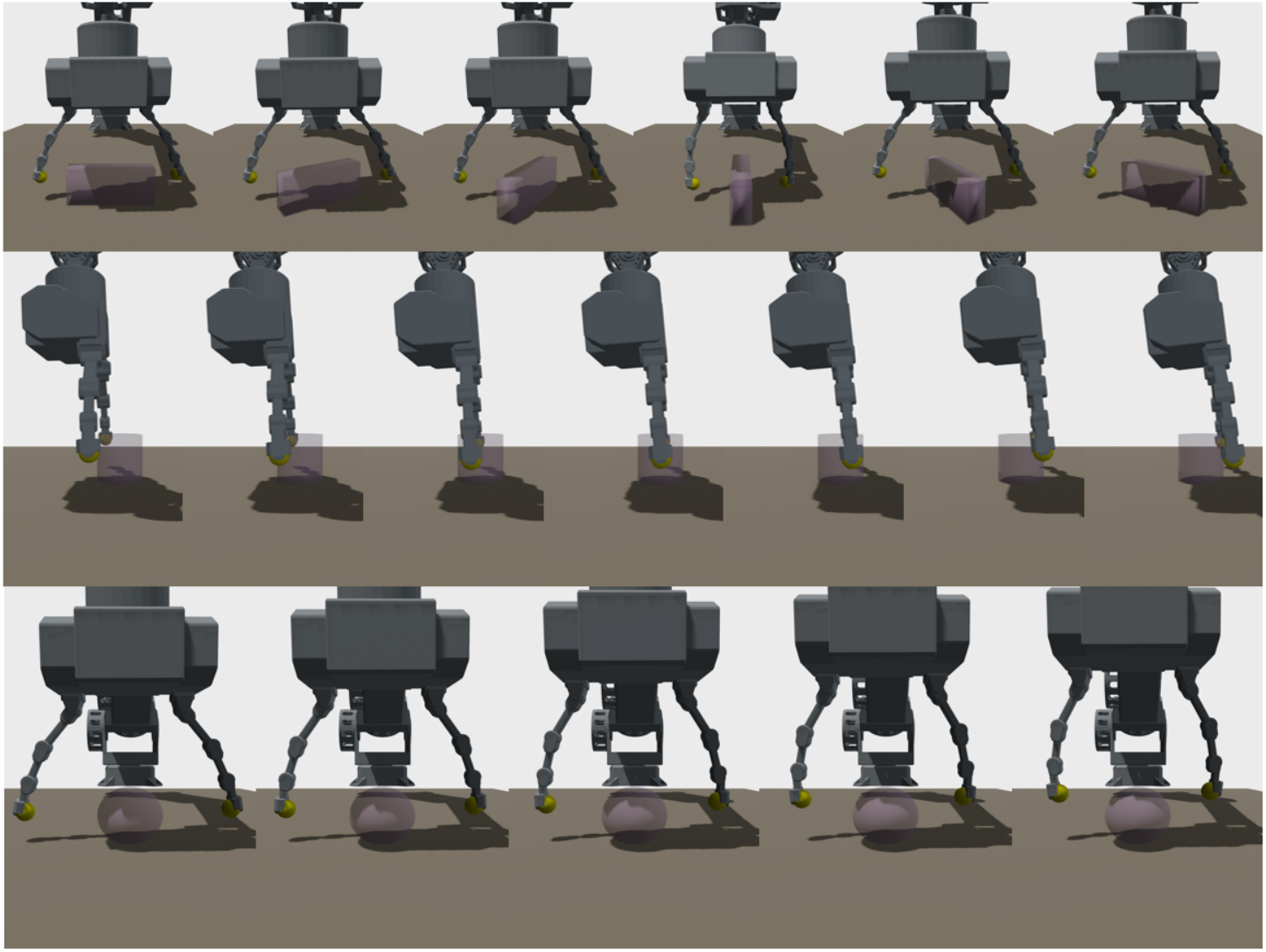}
    \caption{Diverse pre-closure gripper configurations are shown for boxes, cylinders, and ellipsoids.}
    \label{fig:diverse_preclosure}
\end{figure}

To reach a configuration ready for closing the gripper, we employ a simple reactive reaching velocity field defined in the fingertip center coordinates as
\begin{equation}
    \frac{\dot{x}_{1, {\rm des}} + \dot{x}_{2, {\rm des}}}{2} = x^* - (\frac{x_{1} + x_{2}}{2}),
\end{equation}
where $x^*$ is the target position (e.g., the object center),
and solve the QP similar to (\ref{eq:qp_formulation}) to obtain the corresponding joint velocity. 
Specifically, we place the target object below the robot’s default position so that this simple policy brings the gripper to a reasonable pre-closure configuration (Fig.~\ref{fig:diverse_preclosure}).
For the box objects, we applied rotations around the vertical axis (the table surface normal) so that the gripper’s axes were not perfectly aligned with the box at reach. For the cylinder and ellipsoid objects, we introduced a slight offset in the target position relative to the true object center -- mimicking perception errors: for cylinders, the offset was along the horizontal axis in the image plane, and for ellipsoids, along the table surface normal.
\begin{figure}[!t]
    \centering
    \includegraphics[width=\linewidth]{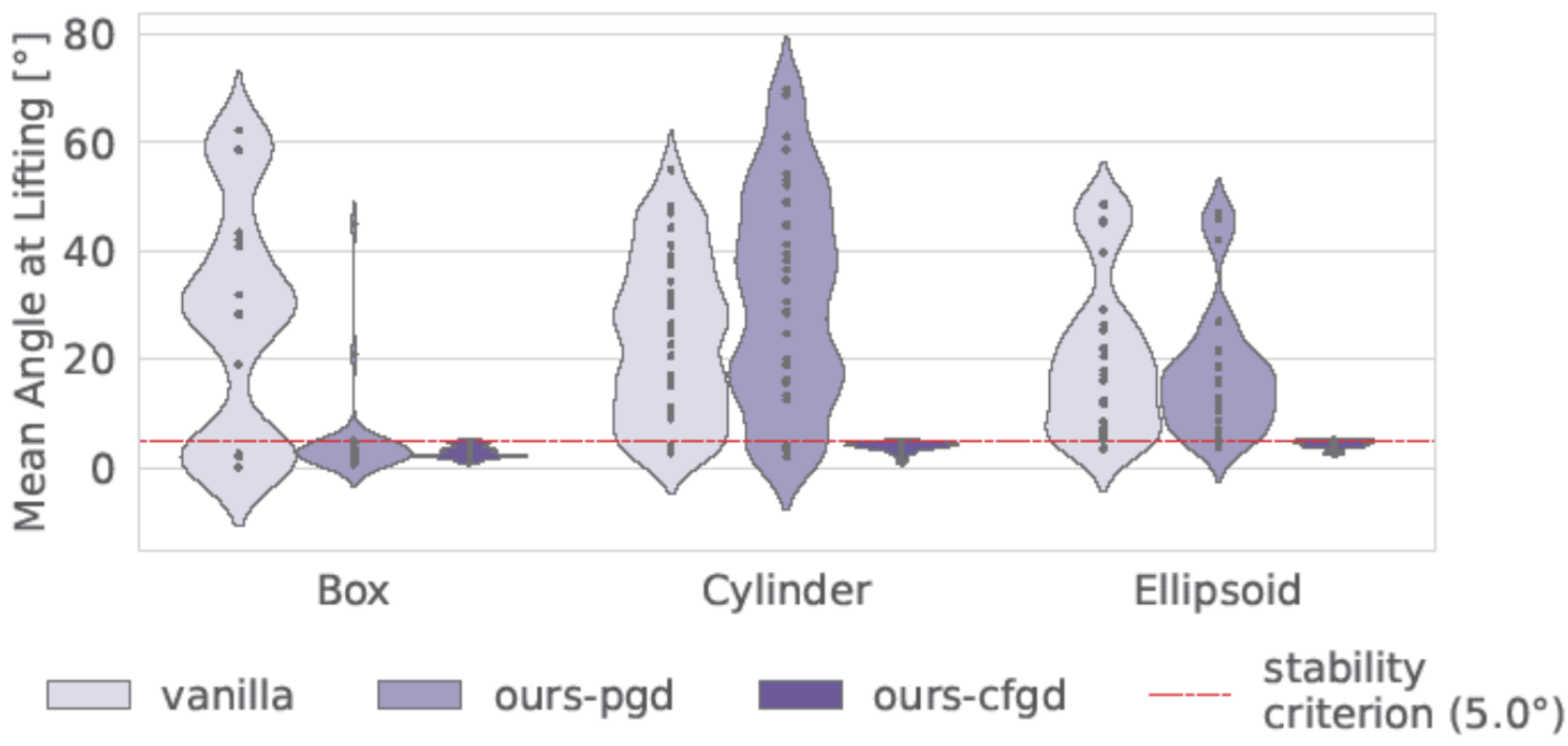}
    \caption{The distribution of the mean angle \(0.5 \, (\phi_1 + \phi_2)\) at lifting is shown, where the vanilla method begins lifting as soon as contacts are detected on both fingertips, while our reflexive grasping methods start lifting only after the stable grasp criterion is satisfied.
}
    \label{fig:exp_results}
\end{figure}

\begin{figure}[!t]
    \centering
    \includegraphics[width=1\linewidth]{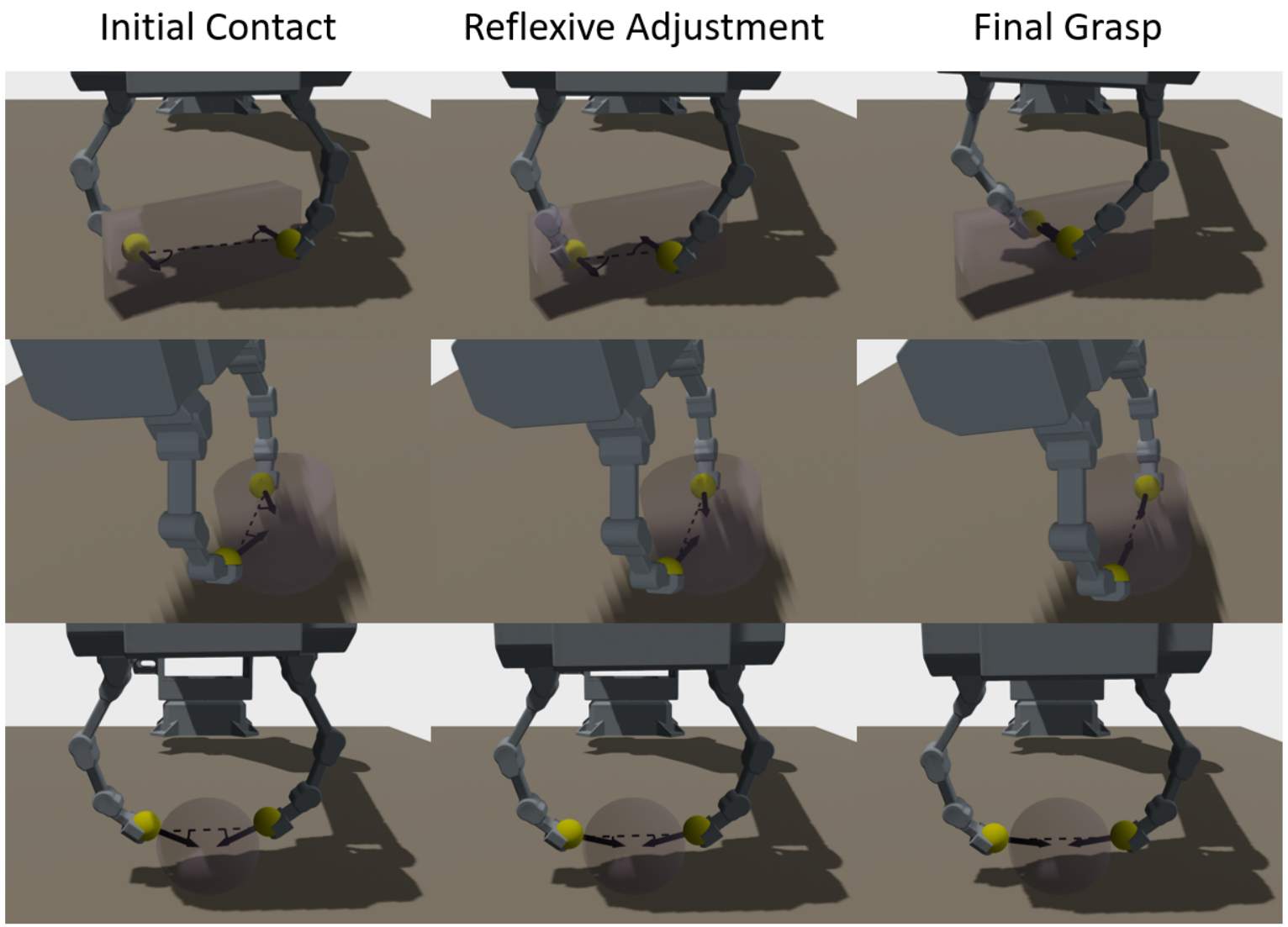}
    \caption{Our reflexive grasping control results with CFGD for box, cylinder, and ellipsoid.}
    \label{fig:exp_vi_results}
\end{figure}

Fig.~\ref{fig:exp_results} shows the distributions of the mean angle \(\frac{\phi_1 + \phi_2}{2}\) at lifting for the vanilla method (no reflex) and our methods (with PGD and CFGD). 
For the vanilla method, once contacts are detected, the gripper attempts to lift the object while applying closing force between the fingertips, without employing the reflexive adjustment algorithm.
The angle distribution of the vanilla method can also be understood as the initial angle distribution for our methods before reflexive grasp adjustment.
Our method with CFGD performs very well across all tested cases. For boxes, PGD is less prone to getting stuck in local minima; however, for cylinders and ellipsoids, PGD does not reduce the angle significantly because the fingertips become trapped in local minima, which is consistent with the results shown in the previous section.
Fig.~\ref{fig:exp_vi_results} shows examples of the results of our reflexive grasp readjustment using CFGD. 

\subsection{Real Robot Experiments}
In this section, we present hardware results of the proposed reflexive grasping algorithm with CFGD by using a 15-DoF arm–hand system (7-DoF arm and 8-DoF hand) adapted from~\cite{saloutos2023towards}, equipped with spherical fingertip tactile sensors~\cite{saloutos2023design}.
The sensors are calibrated using neural networks to estimate contact probabilities and contact positions (under a single-point contact assumption); see~\cite{saloutos2023design} for details. 

One notable difference from our simulation studies is that, in the grasp-adjustment phase, we added a small negative normal velocity component to~(\ref{eq:CFGD}), i.e.,
\begin{equation}
\dot{x}_{i,\mathrm{des}} \;\mapsto\; \dot{x}_{i,\mathrm{des}} - n_i V_n,
\end{equation}
with a small scalar $V_n$. This modification is necessary because the rubber surface is sticky and has a high friction coefficient; thus, purely tangential motions can induce undesired frictional forces on the object and inadvertently alter its position.

\begin{figure}[!t]
    \centering
    \includegraphics[width=\linewidth]{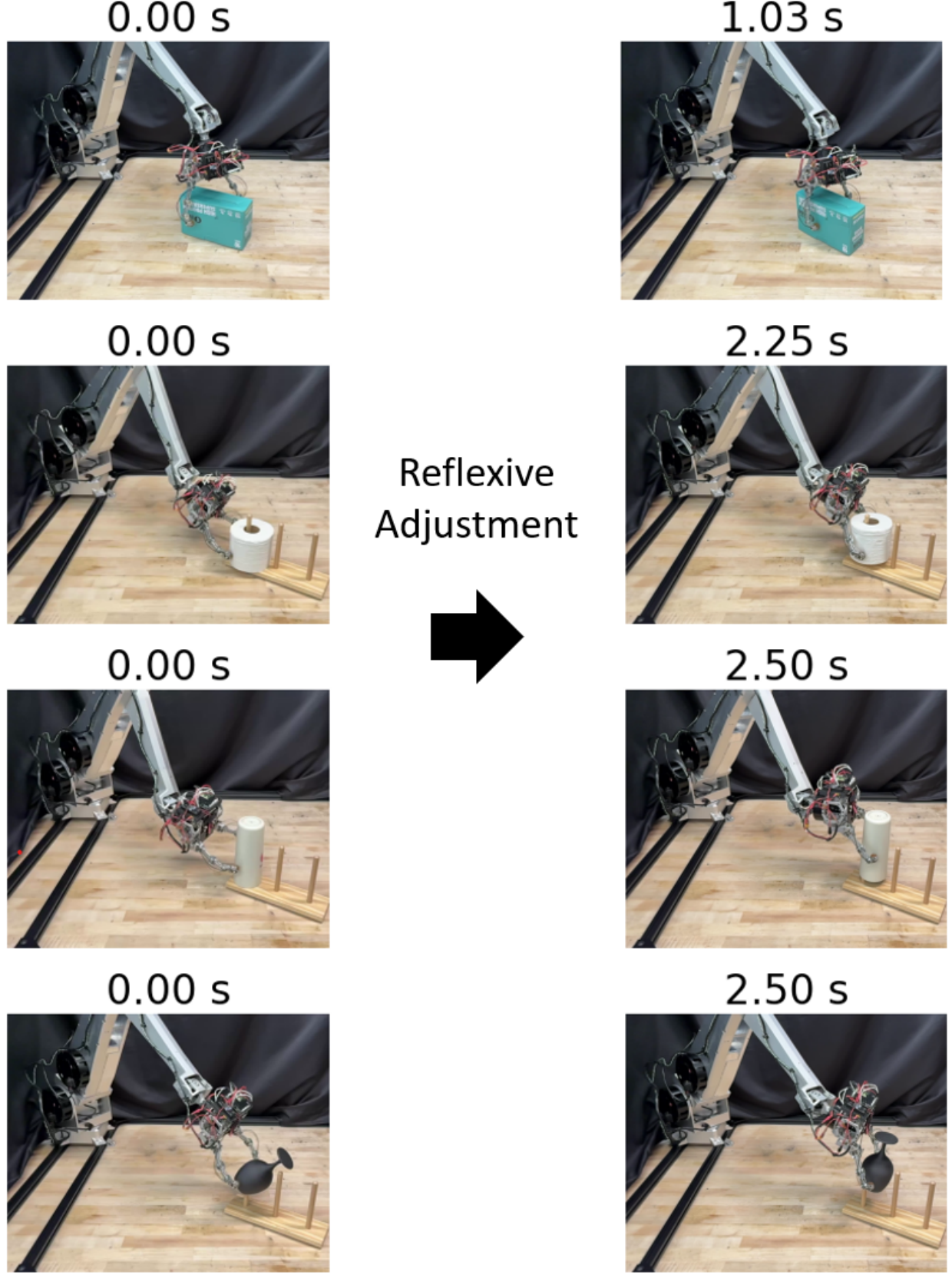}
    \caption{Real-world examples of reflexive grasp adjustment. {\it Left}: initial unstable grasp configuration; {\it Right}: antipodal grasp achieved after reflexive adjustment.}
    \label{fig:real_results}
\end{figure}

Real-world results are shown in Fig.~\ref{fig:real_results}, illustrating the transition from an initial unstable grasp configuration to a final antipodal grasp achieved through reflexive adjustment. The adjustment is completed within one to two seconds, demonstrating the system’s dynamic adaptation capability enabled by high-bandwidth tactile feedback. A slight movement of the box can be observed, as it rotates during the interaction. In other examples, the objects are intentionally placed in an inverted rack. Without this support, the lightweight objects would inevitably be pushed or displaced as the fingertips interact with them to acquire tactile information. This occurs because detecting contact and obtaining reliable contact position estimates requires applying a certain level of normal force -- a limitation of the tactile sensors.

\begin{figure}[!t]
    \centering
    \includegraphics[width=1\linewidth]{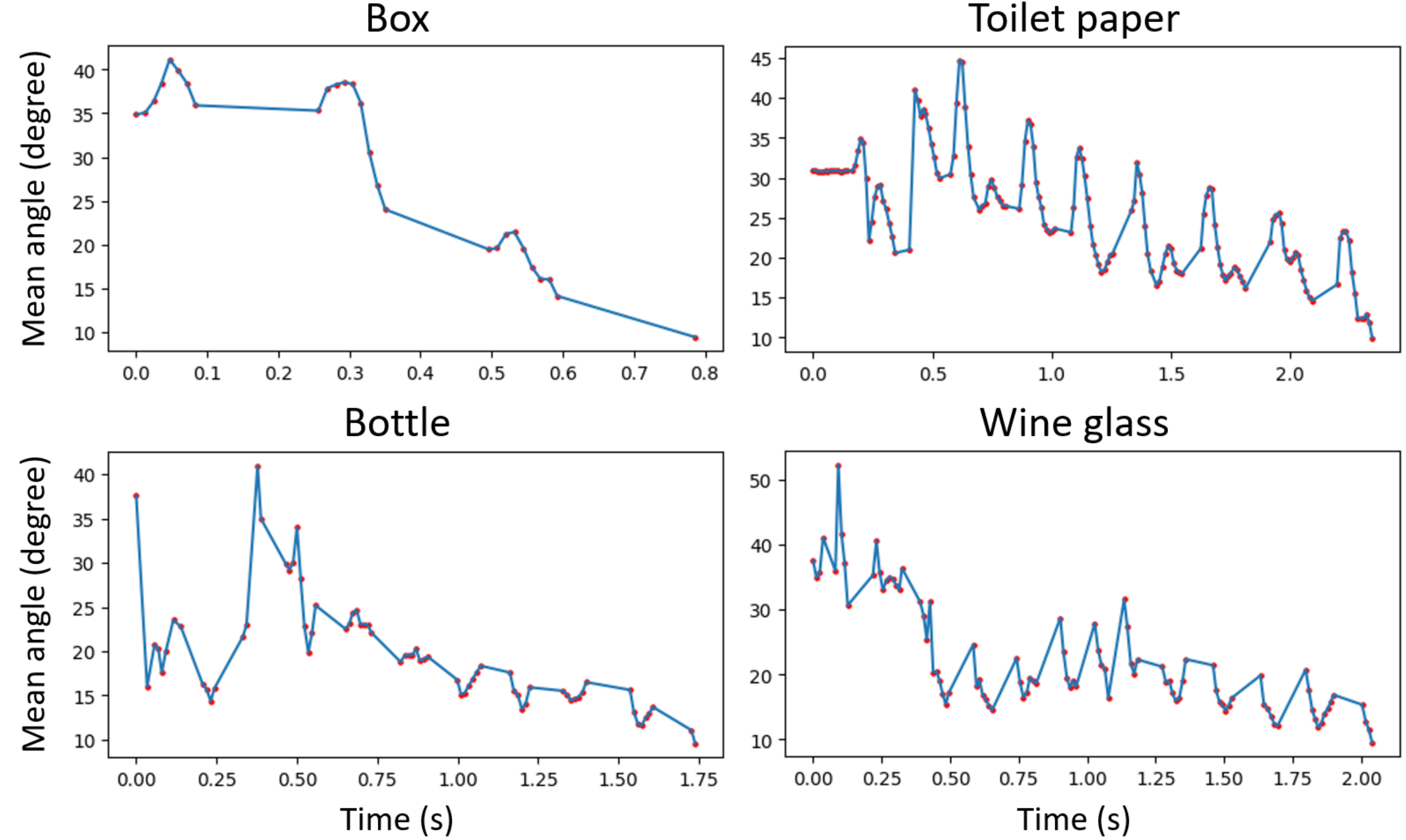}
    \caption{Mean angle $0.5 \; (\phi_1 + \phi_2)$ -- smaller values indicate greater stability -- plotted as a function of time during reflexive grasping. Red dots indicate time steps when both fingertips are in contact.}
    \label{fig:real_graph}
\end{figure}

Fig.~\ref{fig:real_graph} shows the evolution of the mean angle -- defined as half of the grasp stability function $\phi_1 + \phi_2$ in (\ref{eq:antipodal_grasp_stability}) -- over time during reflexive grasping.
Overall, the angles decrease, and once they fall below $10^\circ$ -- the threshold set for the real experiments -- the objects are lifted.
For the toilet paper, and particularly the bottle, the angle does not decrease initially because the object is pushed backward and displaced; once it becomes supported against the rack, the angle begins to decrease. In all cases, high-frequency noise is present, arising from a combination of factors such as tactile sensor noise and control tracking errors.

\section{Discussion and Conclusion}
We have proposed a tactile-reactive grasp readjustment algorithm that relies on contact position information from a high-bandwidth tactile sensor, without making any assumptions about the object’s geometry. Focusing on a two-finger gripper equipped with spherical tactile sensors, we introduced Cross-Finger Gradien`t Descent (CFGD) method, which is empirically verified to converge in nearly all cases for the antipodal grasp stability function. The proposed methods have been validated in both simulation and real-world experiments.

In hardware experiments, the major bottleneck lies in the tactile sensor’s reliability and resolution. With the sensors from~\cite{saloutos2023design}, a critical limitation is the minimum normal force required to obtain a reliable contact point estimate, which necessitated the use of a rack to prevent objects from being pushed away. Note that our algorithm can be applied as long as the tactile sensors provide contact position information along with the corresponding surface normals, and thus can directly benefit from more sensitive tactile sensors in the future.

While the scope of this work is limited to pinch grasps relying solely on fingertip contacts, tactile-reactive grasping methods should ideally be extended to broader configurations, including power grasps that incorporate the palm and contacts along the inner surfaces of the finger links. As the number of contacts increases, the problem complexity grows substantially, making it difficult to establish effective stability criteria and gradient directions; addressing this challenge is an important direction for future research.







\bibliographystyle{IEEEtran}  
\bibliography{ref}

\end{document}